\documentclass{article}

\usepackage{arxiv}

\usepackage[utf8]{inputenc} 
\usepackage[T1]{fontenc}    
\usepackage{hyperref}       
\usepackage{url}            
\usepackage{booktabs}       
\usepackage{amsfonts}       
\usepackage{nicefrac}       
\usepackage{microtype}      
\usepackage{lipsum}
\usepackage{fancyhdr}       
\usepackage{graphicx}       
\graphicspath{{media/}}     
\usepackage{algorithm}
\usepackage{algorithmic}
\usepackage{amsmath}
\usepackage{amssymb}
\usepackage{amsthm}
\newtheorem{definition}{Definition}
\usepackage{mathrsfs}
\usepackage{mathtools}
\newtheorem{Proposition}{Proposition}
\usepackage{enumitem}
\usepackage{booktabs} 
\title{Conditional Uncertainty Quantification for Tensorized Topological Neural Networks }

\author{
  Yujia Wu \\
  Center for Data Science \\
  New York University \\
  \texttt{yw7702@nyu.edu} \\
   \And
  Bo Yang \\
  Department of Biostatistics \\
  University of Michigan - Ann Arbor \\
  \texttt{ybb@umich.edu} \\
   \And
    Yang Zhao \\
  Computer and Information Science \\
  Temple University \\
  \texttt{tur77155@temple.edu} \\
   \And
  Elynn Chen \\
  Stern School of Business \\
  New York University \\
  \texttt{elynn.chen@stern.nyu.edu} \\
   \And
  Yuzhou Chen \\
  Department of Statistics \\
  University of California, Riverside \\
  \texttt{yuzhou.chen@ucr.edu} \\
  \And
  Zheshi Zheng \\
  Department of Biostatistics \\
  University of Michigan - Ann Arbor \\
  \texttt{zszheng@umich.edu} \\
}

\begin{document}
\maketitle

\begin{abstract}
Graph Neural Networks (GNNs) have become the de facto standard for analyzing graph-structured data, leveraging message-passing techniques to capture both structural and node feature information. However, recent studies have raised concerns about the statistical reliability of uncertainty estimates produced by GNNs. This paper addresses this crucial challenge by introducing a novel technique for quantifying uncertainty in non-exchangeable graph-structured data, while simultaneously reducing the size of label prediction sets in graph classification tasks.
We propose Conformalized Tensor-based Topological Neural Networks (CF-T²NN), a new approach for rigorous prediction inference over graphs. CF-T²NN employs tensor decomposition and topological knowledge learning to navigate and interpret the inherent uncertainty in decision-making processes. This method enables a more nuanced understanding and handling of prediction uncertainties, enhancing the reliability and interpretability of neural network outcomes.
Our empirical validation, conducted across 10 real-world datasets, demonstrates the superiority of CF-T²NN over a wide array of state-of-the-art methods on various graph benchmarks. This work not only enhances the GNN framework with robust uncertainty quantification capabilities but also sets a new standard for reliability and precision in graph-structured data analysis.
\end{abstract}


\section{Introduction}
Graph neural networks (GNNs) have emerged as a powerful machinery for various graph learning tasks, such as link prediction, node, and graph classification~\cite{zhou2020graph,xia2021graph}. Their ability to effectively model relational data has transformed our approach to understanding complex systems, including social networks \cite{li2021graph}, biochemical structures \cite{chen2018survey}, traffic networks \cite{qian2023trafficuncertainty}.
As GNNs increasingly play a role in critical decision-making in social and medical applications, the accuracy of their predictions and their reliability have become paramount. In particular, the uncertainty quantification of label prediction is of great interest in practice. For example, a predicted probability of $\{0.49,0.51\}$ and $\{0.1,0.9\}$ for labels A and B will both lead to predicted label B, but the confidence level will be quite different for the two cases. Thus, an increasing number of researchers are focusing on the uncertainty of predictions made by GNNs~\cite{hsu2022makes,zhang2020mix,lakshminarayanan2017simple,wang2021confident}. A recent survey \cite{wang2024uncertainty} summarizes several uncertainty quantification approaches, including Bayesian-based methods. Nonetheless, many techniques frequently fall short in providing both theoretical and empirical assurances about their accuracy, specifically the likelihood that the predicted set or interval encompasses the actual outcome~\cite{angelopoulos2020uncertainty}. This deficiency in precision impedes their trustworthy application in contexts where mistakes may have significant repercussions.

Conformal prediction~\cite{vovk2005algorithmic} (CP) is a framework for producing statistically guaranteed prediction with uncertainty quantification. The general idea is to assess the conformity level of the potential value of outcome with respect to the observed data. So conformity, or in other words, ``{\em exchangeability}'', is a crucial assumption for this method.
This assumption is easily violated if there are covariate shifts or covariate constraints in the graph features which is commonly observed in large data sets. Here covariate shifts mean the distribution of the covariates is different, and covariate constraints mean the target of prediction is in a single or sub-type of graph features.
Most works in GNN uncertainty quantification do not deal with those issues \cite{huang2023uncertainty,zargarbashi2023conformal}. Existing works focusing on non-exchangeable cases are often developed by adjusting the conformal prediction interval according to the estimated covariate shift \cite{tibshirani2019conformal,gibbs2021adaptive}. 
However, they are not directly applicable to GNNs where the shifts of graph covariates are hard to estimate. They are not applicable to the case where the target is to quantify the uncertainty of a single or certain type of graph.

To address this challenge, we develop a novel algorithm that produces a statistically coverage-guaranteed conditional prediction set under possible graph covariate shifts. 
This set not only offers guaranteed coverage under a relaxed condition of exchangeability, accommodating potential domain shifts in graph data but also innovates by requiring the existence of a local calibration set that contains ``similar graph data" as the target. Termed the "local exchangeable" condition, this approach significantly enhances the adaptability of prediction models.

The primary obstacle in employing this conditional prediction set lies in the construction of the local calibration set, a task complicated by the high-dimensional and multi-dimensional nature of graph data. To navigate this complexity, we have developed a methodology that leverages Tensorized Topological Neural Networks, crafting our similarity measures on the basis of tensorized graph structure and its low-rank tensor decompositions. This approach effectively captures the essence of multi-dimensionality while simultaneously reducing dimensionality, thus facilitating a more nuanced and accurate modeling of graph data.

Through rigorous empirical validation across ten real-world datasets, our proposed methodologies have demonstrated superior performance against a comprehensive range of leading-edge methods in various graph analysis benchmarks. This accomplishment not only strengthens the GNN framework with robust uncertainty quantification tools but also establishes a new benchmark for reliability and precision in the analysis of graph-structured data.
\section{Related Work}
\paragraph{\bf Uncertainty Quantification.}
Conformal prediction is an attractive framework for uncertainty quantification of GNN lable prediction. 
\cite{huang2023uncertainty} and \cite{zargarbashi2023conformal} provide prediction set with guaranteed coverage and controled sizes. However, \cite{zargarbashi2023conformal} utilizes the propagation of node labels in the calibration set, whereas \cite{huang2023uncertainty} employs the inefficiency of the calibration set itself as the criterion for optimization. 
Several procedures of conformal prediction are introduced in \cite{lei2018distribution}, including split conformal we adopted in this article, which is both theoretically and computationally easy. \cite{xie2022homeostasis} provides a detailed discussion of its properties and introduces a new concept of conditional prediction inference that challenges the coverage guarantee for conformal prediction. Also shown in \cite{foygel2021limits}, conditional inference is impossible without additional assumptions. Thus we develop an algorithm to handle this issue by adopting the so-called ``iFusion learning'' idea \cite{shen2020fusion,cai2023individualized}: instead of using the randomly selected calibration set to quantify uncertainty, we use a subset that contains only ``similar'' items to make inference for our target. 

\paragraph{\bf  Tensor Learning.}
Tensors, representing multidimensional data, have become prominent across various scientific disciplines, including neuroimaging \cite{zhou2013tensor}, economics \cite{chen2020constrained,liu2022identification}, international trade \cite{chen2022modeling}, recommendation systems \cite{bi2018multilayer}, multivariate spatial-temporal data analysis \cite{chen2020modeling}, and biomedical applications \cite{chen2024semi,chen2024distributed}. Key areas of tensor research encompass tensor decomposition \cite{zhang2018tensor,chen2024distributed}, tensor regression \cite{zhou2013tensor,li2017parsimonious,XiaZhangZhou2022,chen2024factor}, and unsupervised learning techniques such as tensor clustering \cite{SunLi2019,mai2021doubly,LuoZhang2022}.

Empirical studies have demonstrated the effectiveness of tensor-based classification methods across these fields, leading to the development of techniques including support tensor machines \cite{hao2013linear, guoxian2016}, tensor discriminant analysis \cite{chen2024high}, tensor logistic regression \cite{wimalawarne2016theoretical}, and tensor neural networks \cite{kossaifi2020tensor,wen2024tensor}.
Neural networks designed to process tensor inputs enable efficient analysis of high-dimensional data. However, most existing work employs tensors primarily for computation rather than statistical analysis. Notable advancements include the analysis of deep neural networks' expressive power using tensor methods \cite{cohen2016expressive}, which established an equivalence between neural networks and hierarchical tensor factorizations. The Tensor Contraction Layer (TCL) \cite{kossaifi2017tcl} introduced tensor contractions as trainable neural network layers, while the Tensor Regression Layer (TRL) \cite{kossaifi2020trl} further regularized networks through regression weights. The Graph Tensor Network (GTN) \cite{xu2023graph} proposed a Tensor Network-based framework for describing neural networks using tensor mathematics and graphs for large, multi-dimensional data.

Despite these advancements, current neural networks with tensor inputs lack comprehensive theoretical and empirical exploration. The Tensor-view Topological Graph Neural Network (TTG-NN) \cite{wen2024tensor} represents a step forward, introducing a novel class of topological deep learning built upon persistent homology, graph convolution, and tensor operations.

\paragraph{\bf Persistent Homology.}
Persistent homology (PH)~\cite{edelsbrunner2000topological,zomorodian2005computing} is a suite of tools within TDA that has shown great promise in a broad range of domains including bioinformatics, material sciences, and social networks~\cite{otter2017roadmap,aktas2019persistence}. One of the key benefits of PH is that it can capture subtle patterns in the data shape dynamics at multiple resolution scales. PH has been successfully integrated as a fully trainable topological layer into various machine learning and deep learning models~\cite{pun2018persistent}, addressing such tasks as image classification~\cite{hofer2017deep}, 2D/3D shape classification~\cite{bonis2016persistence}, molecules and biomolecular complexes representation learning~\cite{cang2018representability}, graph classification~\cite{chen2023topological}, and spatio-temporal prediction~\cite{chen2021z}. For example, ~\cite{carriere2019perslay} builds a neural network based on the DeepSet architecture~\cite{zaheer2017deep} which can achieve end-to-end learning for topological features. \cite{cang2018representability} introduces multi-component persistent homology, multi-level persistent homology and electrostatic persistence for chemical and biological characterization, analysis, and modeling by using convolutional neural networks. \cite{horntopological} proposes a trainable topological layer that incorporates global topological information of a graph using persistent homology. ~\cite{chen2022time} proposes a time-aware topological deep learning model that captures interactions and encodes encodes time-conditioned topological information. However, to the best of our knowledge, both TDA and PH have yet to be employed for uncertainty quantification.
\section{Preliminaries}
\paragraph{\bf Problem Settings.} 
Let $\mathcal{G} = (\mathcal{V}, \mathcal{E}, \boldsymbol{X})$ be an attributed graph, where $\mathcal{V}$ is a set of nodes ($|\mathcal{V}|=N$), $\mathcal{E}$ is a set of edges, and $\boldsymbol{X} \in \mathbb{R}^{N \times F}$ is a feature matrix of nodes (here $F$ is the dimension of the node features). 
Let $\boldsymbol{A} \in \mathbb{R}^{N \times N}$ be a symmetric adjacency matrix whose entries are
$a_{ij} = \omega_{ij}$ if nodes $i$ and $j$ are connected and 0 otherwise 
(here $\omega_{ij}$ is an edge weight and $\omega_{ij}\equiv 1$ for unweighted graphs).
Furthermore, $\mathbf{D}$ represents the degree matrix of $\boldsymbol{A}$, that is $d_{ii} = \sum_j a_{ij}$. 
In the graph classification setting, we have a set of graphs $\{\mathcal{G}_1, \mathcal{G}_2, \dots, \mathcal{G}_\aleph\}$, where each graph $\mathcal{G}_i$ is associated with a label $y_i$. 
We denote a sample consisting of a graph and its label as $\overline{\mathcal{G}}_i = (\mathcal{G}_i, y_i)$. 
The goal of the graph classification task is to take the graph as the input and predict its corresponding label. Given a graph $\mathcal{G}$, the trained model $\widehat{f}(\mathcal{G})$ has a predicted label $\hat{y}$, which is the $argmax$ of the probability distribution over all possible labels, say $\{1,\dots,k\}$. 
To apply the uncertainty quantification algorithm, our goal is to obtain a prediction set $C(\mathcal{G}) = C(\mathcal{G},\alpha)$ with a certain level of confidence $\alpha\in(0,1)$ for label $y$. We split our dataset $\{\overline{\mathcal{G}}_1, \overline{\mathcal{G}}_2, \dots, \overline{\mathcal{G}}_\aleph\}$ into four disjoint subset, $\overline{\mathcal{G}}_{\text{train}}$, $\overline{\mathcal{G}}_{\text{valid}}$, $\overline{\mathcal{G}}_{\text{calib}}$, $\overline{\mathcal{G}}_{\text{test}}$, of fixed size with pre-defined ratios. They correspond to training, validation, calibration, and test sets. The calibration set $\overline{\mathcal{G}}_{\text{calib}}$ is withheld to apply conformal prediction later for uncertainty quantification. 

\paragraph{\bf Persistent Homology.} PH is a subfield in computational topology whose main goal is to detect, track and encode the evolution of shape patterns in the observed object along various user-selected geometric dimensions~\cite{edelsbrunner2000topological,zomorodian2005computing,carlsson2021topological}. These shape patterns represent topological properties such as connected components, loops, and, in general, $n$-dimensional "holes", that is, the characteristics of the graph $\mathcal{G}$ that remain preserved at different resolutions under continuous transformations. By employing such a multi-resolution approach, PH addresses the intrinsic limitations of classical homology and allows for retrieving the latent shape properties of $\mathcal{G}$ which may play an essential role in a given learning task. The key approach here is to select some suitable scale parameters $\nu$ and then to study changes in the shape of $\mathcal{G}$ that occur as $\mathcal{G}$ evolves with respect to $\nu$. That is, we no longer study $\mathcal{G}$ as a single object but as a {\it filtration} $\mathcal{G}_{\nu_1} \subseteq \ldots \subseteq \mathcal{G}_{\nu_n}=\mathcal{G}$, induced by monotonic changes of $\nu$. To ensure that the process of pattern selection and counting is objective and efficient, we build an abstract simplicial complex $\mathscr{K}(\mathcal{G}_{\nu_j})$ on each $\mathcal{G}_{\nu_j}$, which results in filtration of complexes $\mathscr{K}(\mathcal{G}_{\nu_1}) \subseteq \ldots \subseteq \mathscr{K}(\mathcal{G}_{\nu_n})$. 
For example, for an edge-weighted graph $(\mathcal{V}, \mathcal{E}, w)$, with the edge-weight function $w: \mathcal{E} \rightarrow \mathbb{R}$, we can set $\mathcal{G}_{\leq\nu_j}=(\mathcal{V}, \mathcal{E}, w^{-1}(-\infty, \nu_j])$ for each $\nu_j$, $j=1,\ldots, n$, yielding the induced sublevel edge-weighted filtration. 
Similarly, we can consider a function on a node set $\mathcal{V}$, for example, node degree, which results in a sequence of induced subgraphs of $\mathcal{G}$ with a maximal degree of $\nu_j$ for each $j=1,\ldots, n$ and the associated degree sublevel set filtration. We can then record scales $b_i$ (birth) and $d_i$ (death) at which each topological feature first and last appear in the sublevel filtration
$\mathcal{G}_{\nu_1} \subseteq \mathcal{G}_{\nu_2} \subseteq \mathcal{G}_{\nu_3} \ldots \subseteq \mathcal{G}_{\nu_n}$. In this paper, to encode the above topological information presented in a PD ${Dg}$ into the embedding function, we use its vectorized representation, i.e., persistence image (PI)~\cite{adams2017persistence}. The PI is a finite-dimensional vector representation obtained through a weighted kernel density function and can be computed in the following two steps (see more details in Definition~\ref{def_pi}). First, we map the PD ${Dg}$ to an integrable function $\varrho_{{Dg}}: \mathbb{R}^{2} \mapsto \mathbb{R}^{2}$, which is referred to as a persistence surface. The persistence surface $\varrho_{{Dg}}$ is constructed by summing weighted Gaussian kernels centered at each point in ${Dg}$. In the second step, we integrate the persistence surface $\varrho_{{Dg}}$ over each grid box to obtain the value of the ${PI}_{Dg}$. 

    \begin{definition}[Persistence Image]\label{def_pi}
    Let $g: \mathbb{R}^2 \mapsto \mathbb{R}$ be a non-negative weight function for the persistence plane $\mathbb{R}$. The value of each pixel $z \in \mathbb{R}^2$ is defined as ${PI}_{{Dg}}(z) = \iint\limits_{z} \sum_{\mu \in T({Dg})} \frac{g(\mu)}{2 \pi \delta_{x} \delta_{y}} e^{-\left(\frac{\left(x-\mu_{x}\right)^2}{2\delta_{x}^{2}}+\frac{\left(y-\mu_{y}\right)^2}{2\delta_{y}^{2}}\right)} d y d x,$ where $T(\text{Dg})$ is the transformation of the PD $\text{Dg}$ (i.e., for each $(x,y)$, $T(x,y) = (x, y-x)$), $\mu = (\mu_x, \mu_y) \in \mathbb{R}^2$, and $\delta_x$ and $\delta_y$ are the standard deviations of a differentiable probability distribution in the $x$ and $y$ directions respectively. 
    \end{definition}

\paragraph{\bf Conformal Prediction and Its Limitations.}
Conformal prediction \cite{lei2018distribution} typically imposes the ``exchangeable'' condition on $\overline{\mathcal{G}}_{valid}$, $\overline{\mathcal{G}}_{calib}$, and $\overline{\mathcal{G}}_{test}$, that is
\begin{equation}\label{eq:exchangeable}
\overline{\mathcal{G}}_{valid} \sim \overline{\mathcal{G}}_{calib} \sim \overline{\mathcal{G}}_{test} \overset{i.i.d}{\sim} \overline{\mathcal{F}},  
\end{equation}
where $\overline{\mathcal{F}}$ is the joint distribution of $\mathcal{G}$ and its label $y$.

The goal of conformal prediction is to construct a prediction set $C(\mathcal{G})$ of a predicted label $y$ for any graph $\mathcal{G}\in\mathcal{G}_{test}$, such that the unconditional coverage 
\begin{equation}\label{eq: cov uncond}
\mathbb{P}(y \in C(\mathcal{G})) \geq 1 - \alpha,
\end{equation}
for any user-chosen error rate $\alpha\in[0,1]$, and the probability measure is defined for the joint data $\overline{\mathcal{G}}_{train}\cup\overline{\mathcal{G}}_{calib}\cup(\mathcal{G},y)$. 

The ``{\em exchangeable}'' condition \eqref{eq:exchangeable} is restrictive and can be violated in many applications. For example, we collect data and train the algorithm on a large population, but are only interested in applying the algorithm to a specific sub-population and $\mathcal{G}_{test}$ and $\mathcal{G}_{valid}$ are non-exchangeable with $\mathcal{G}_{calib}$ and $\mathcal{G}_{train}$.
In this paper, we relax this assumption extensively to allow for possible domain shifts/constrain in graph data sets. 

\begin{figure*}[ht]
  \centering
  \includegraphics[width=\linewidth]{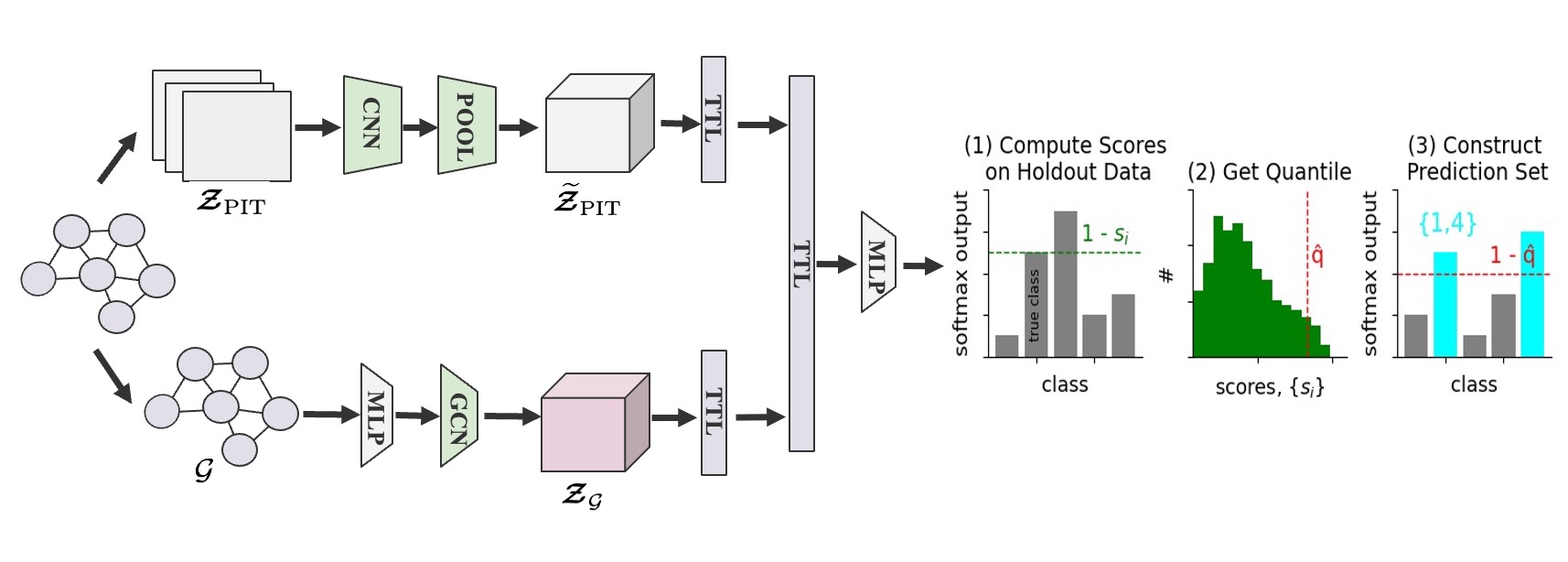}
  \caption{The architecture of CF-T$^2$NN.}
  \label{fig:flowchart}
\end{figure*}
\section{Method}
\paragraph{\bf Tensor Transformation Layer.}
The {\em Tensor Transformation Layer (TTL)} preserves the tensor structures of feature $\boldsymbol{\boldsymbol{\mathcal{X}}}$ of dimension $D=\prod_{m=1}^M D_m$ and hidden throughput.
Let $L$ be any positive integer 
and $\mathbf{d} = \left[d^{(1)}, \cdots, d^{(L+1)}\right]$ collects the width of all layers.

A {\em deep ReLU Tensor Neural Network} is a function mapping taking the form of
\begin{equation} \label{eqn:TNN}
f(\boldsymbol{\mathcal{X}}) = \mathcal{L}^{(L+1)} \circ \sigma \circ \mathcal{L}^{(L)} \circ \sigma \cdots \circ \mathcal{L}^{(2)} \circ \sigma \circ \mathcal{L}^{(1)}(\boldsymbol{\mathcal{X}})
\end{equation}
where $\sigma(\cdot)$ is an element-wise activation function. 
Affine transformation $\mathcal{L}^{(\ell)}(\cdot)$ and the hidden input and output tensor of the $\ell$-th layer, i.e., $\boldsymbol{\mathcal{H}}^{(\ell+1)}$ and $\boldsymbol{\mathcal{H}}^{(\ell)}$, are defined by
\begin{equation} \label{trl_eq}
\begin{aligned}
\mathcal{L}^{(\ell)}\left(\boldsymbol{\mathcal{H}}^{(\ell)}\right) \coloneqq \langle \boldsymbol{\mathcal{W}}^{(\ell)}, \boldsymbol{\mathcal{H}}^{(\ell)} \rangle + \boldsymbol{\mathcal{B}}^{(\ell)}, \\
\text{and}\quad
\mathbf{\mathcal{H}}^{(\ell+1)} \coloneqq \sigma\left(\mathcal{L}^{(\ell)}\left(\mathbf{\mathcal{H}}^{(\ell)}\right)\right)
\end{aligned}
\end{equation}
where
$\boldsymbol{\mathcal{H}}^{(0)}= \boldsymbol{\mathcal{X}}$ takes the tensor feature, 
$\langle \cdot, \cdot \rangle$ is the tensor inner product, 
and {\em low-rank weight} tensor $\boldsymbol{\mathcal{W}}^{(\ell)}$ and a bias tensor $\boldsymbol{\mathcal{B}}^{(\ell))}$. 
The tensor structure kicks in when we incorporate tensor low-rank structures such as {\it CP low-rank}, {\it Tucker low-rank}, and {\it Tensor Train low-rank}.

Tucker low-rank structure is defined by
\begin{equation} \label{eqn:tucker}
    \boldsymbol{\mathcal{X}} = \boldsymbol{\mathcal{C}} \times_1 \mathbf{U}_1 \times_2 \cdots \times_M \mathbf{U}_M + \boldsymbol{\mathcal{E}},
\end{equation}
where 
$\boldsymbol{\mathcal{E}} \in \mathbb{R}^{D_1 \times \cdots \times D_M}$ is the tensor of the idiosyncratic component (or noise) and 
$\boldsymbol{\mathcal{C}}$is the latent core tensor representing the true low-rank feature tensors, and $\mathbf{U}_m, \, m \in [M]$, are the loading matrices.  

The complete definitions of three low-rank structures are given as follows. Consider an $M$-th order tensor $\boldsymbol{\mathcal{X}}$ of dimension $D_1\times\cdots\times D_M$. 
If $\boldsymbol{\mathcal{X}}$ assumes a (canonical) rank-$R$ 
{\em CP low-rank} structure, then it can be expressed as
\begin{equation} \label{eqn:cp}
    \boldsymbol{\mathcal{X}} = \sum_{r=1}^R c_r \, \mathbf{u}_{1r} \circ \mathbf{u}_{2r} \circ \cdots \circ \mathbf{u}_{Mr},
\end{equation}
where $\circ$ denotes the outer product, $\mathbf{u}_{mr} \in \mathbb{R}^{D_m}$ and $\mathbf{u}_{mr}\|_2 = 1$ for all mode $m \in [M]$ and latent dimension $r \in [R]$.
Concatenating all $R$ vectors corresponding to a mode $m$, we have
$\mathbf{u}_{m} = [\mathbf{u}_{m1}, \cdots, \mathbf{u}_{mR}] \in \mathbb{R}^{D_m \times R}$ which is referred to as the loading matrix for mode $m \in [M]$. 

If $\boldsymbol{\mathcal{X}}$ assumes a rank-$(R_1, \cdots, R_M)$  
{\em Tucker low-rank} structure \eqref{eqn:tucker}, then it writes 
\begin{align} 
    \boldsymbol{\mathcal{X}} &= \boldsymbol{\mathcal{C}} \times_1 \mathbf{U}_1 \times_2 \cdots \times_M \mathbf{U}_M \notag = \sum_{r_1=1}^{R_1} \cdots \sum_{r_M=1}^{R_M} c_{r_1 \cdots r_M} 
    \left( \mathbf{u}_{1 r_1} \circ \cdots \circ \mathbf{u}_{M r_M} \right),
\end{align}
where $\mathbf{u}_{m r_m}$ are all $D_m$-dimensional vectors, and $c_{r_1\cdots r_M}$ are elements in the $R_1 \times \cdots \times R_M$-dimensional core tensor $\boldsymbol{\mathcal{C}}$.

\emph{Tensor Train (TT) low-rank} \cite{oseledets2011tensor} approximates a $D_1\times\cdots\times D_M$ tensor $\boldsymbol{\mathcal{X}}$ with a chain of products of third-order \emph{core tensors} $\boldsymbol{\mathcal{C}}_i$, $i\in [M]$, of dimension $R_{i-1}\times D_i\times R_i$.  
Specifically, each element of tensor $\boldsymbol{\mathcal{X}}$ can be written as 
\begin{equation} \label{eqn:tt}
    x_{i_1, \cdots, i_M} = \mathbf{c}_{1,1,i_1,:}^\top
    \times\mathbf{c}_{2,:,i_2,:}\times\cdots\times\mathbf{c}_{M,:,i_M,:}
    \times\mathbf{c}_{M+1,:,1,1}, 
\end{equation}
where $\mathbf{c}_{m,:,i_m,:}$ is an $R_{m-1}\times R_m$ matrix for $m\in[M]\cup\{\text{M+1}\}$.
The product of those matrices is a matrix of size $R_0 \times R_{M+1}$.
Letting $R_0 = 1$, the first core tensor $\mathcal{C}_1$ is of dimension $1 \times D_1 \times R_1$, which is actually a matrix and whose $i_1$-th slice of the middle dimension (i.e. $\mathcal{C}_{1,1,i_1,:}$) is actually a $R_1$ vector. 
To deal with the ``boundary condition'' at the end, we augmented the chain with an additional tensor $\boldsymbol{\mathcal{C}}_{M+1}$ with $D_{M+1} = 1$ and $R_{M+1} = 1$ of dimension $R_M \times 1 \times 1$.  
So the last tensor can be treated as a vector of dimension $R_M$. 

CP low-rank \eqref{eqn:cp} is a special case where the core tensor $\boldsymbol{\mathcal{C}}$ has the same dimensions over all modes, that is $R_m = R$ for all $m\in[M]$, and is super-diagonal. 
TT low-rank is a different kind of low-rank structure and it inherits advantages from both CP and Tucker decomposition.
Specifically, TT decomposition can compress tensors as significantly as CP decomposition, while its calculation is as stable as Tucker decomposition. 

\paragraph{\bf Multi-View Topological Convolutional Layer.}
To capture the underlying topological features of a graph $\mathcal{G}$,  
we employ $\mathcal{K}$ filtration functions: $f_i: \mathcal{V} \mapsto \mathbb{R}$ for $i = \{1, \dots, \mathcal{K}\}$. Each filtration function $f_i$ gradually reveals one specific topological structure at different levels of connectivity, such as the number of relations of a node (i.e., degree centrality score), node flow information (i.e., betweenness centrality score), information spread capability (i.e., closeness centrality score), and other node centrality measurements. With each filtration function $f_i$, we construct a set of $Q$ persistence images of resolution $P \times P$ using tools in persistent homology analysis.

Combining $Q$ persistence images of resolution $P\times P$ from $K$ different filtration functions, we construct a {\it multi-view} topological representation, namely {\em Persistent Image (PI) Tensor} $\boldsymbol{\mathcal{Z}}_\text{PIT}$ of dimension $K \times Q \times P \times P$. We design the {\em Multi-View Topological Convolutional Layer (MV-TCL)} to (i) jointly extract and learn the latent topological features contained in the $\boldsymbol{\mathcal{Z}}_\text{PIT}$, (ii) leverage and preserve the multi-modal structure in the $\boldsymbol{\mathcal{Z}}_\text{PIT}$, and (iii) capture the structure in trainable weights (with fewer parameters). Firstly, hidden representations of the PI tensor $\boldsymbol{\mathcal{Z}}_\text{PIT}$ are achieved through a combination of a CNN-based model and global pooling. 
Mathematically, we obtain a learnable topological tensor representation defined by
\begin{equation}\label{gcn_tda_part}
\boldsymbol{\Tilde{\mathcal{Z}}}_{\text{PIT}} =
    \begin{cases}
    f_{\text{CNN}}(\boldsymbol{\mathcal{Z}}_\text{PIT}) \quad \text{if} \; |Q|=1\\
    \xi_{\text{POOL}}(f_{\text{CNN}}(\boldsymbol{\mathcal{Z}}_\text{PIT})) \quad \text{if} \; |Q| > 1
    \end{cases},
\end{equation}
where $f_{\text{CNN}}$ is a CNN-based neural network, 
$\xi_{\text{POOL}}$ is a pooling layer that preserves the information of the input in a fixed-size representation (in general, we consider either global average pooling or global max pooling). 
Equation \eqref{gcn_tda_part} provides two simple yet effective models to extract {\it learnable} topological features: (i) if only considering $q$-dimensional topological features in $\boldsymbol{{\boldsymbol{\mathcal{Z}}}}_{\text{PIT}}$, we can apply any CNN-based model to learn the latent feature of the $\boldsymbol{{\boldsymbol{\mathcal{Z}}}}_{\text{PIT}}$; 
(ii) if considering topological features with $Q$ dimensions, we can additionally employ a global pooling layer over the latent feature and obtain an image-level feature. In this paper, we treat the resulting topological features in dimension 0 (connected components) and 1 (cycles) (i.e., $\mathcal{Q} = 2$).

\paragraph{\bf Graph Convolutional Layers.}
Our third representation learning module is the Graph Convolutional Layer (GCL). It utilizes the graph structure of ${\mathcal{G}}$ with its node feature matrix $\boldsymbol{X}$ through the graph convolution operation and a multi-layer perceptron (MLP). 
Specifically, the designed graph convolution operation proceeds by multiplying the input of each layer with the $\tau$-th power of the normalized adjacency matrix. 
The $\tau$-th power operator contains statistics from the $\tau$-th step of a random walk on the graph (in this study, we set $\tau$ to be 2), thus nodes can indirectly receive more information from farther nodes in the graph.
Combined with a multi-layer perceptron (MLP), the representation learned at the $\ell$-th layer is given by
\begin{align}
\label{GNN_MLP_eq}
{\boldsymbol{\mathcal{Z}}_{\mathcal{G}}^{(\ell +1)}} = f_{\text{MLP}}({\sigma({\hat{\boldsymbol{A}}^{\tau}\boldsymbol{H}_{\mathcal{G}}^{(\ell)}\boldsymbol{\Theta}^{(\ell)}}})),
\end{align}
where $\hat{\boldsymbol{A}} = \boldsymbol{\Tilde{{D}}}^{-\frac{1}{2}}\boldsymbol{\Tilde{{A}}}\boldsymbol{\Tilde{{D}}}^{\frac{1}{2}}$, 
$\boldsymbol{\Tilde{{A}}} = \boldsymbol{A} + \boldsymbol{I}$, 
and $\boldsymbol{\Tilde{{D}}}$ is the corresponding degree matrix of $\boldsymbol{\Tilde{{A}}}$, $\boldsymbol{H}_{\mathcal{G}}^{(0)} = \boldsymbol{X}$, $f_{\text{MLP}}$ is an MLP which has 2 layers with batch normalization, $\sigma(\cdot)$ is the non-linear activation function, $\boldsymbol{\Theta}^{(\ell)}$ is a trainable weight of $\ell$-th layer. Note that, to exploit multi-hop propagation information and increase efficiency, we can apply tensor decomposition (TD) techniques (e.g., tucker decomposition and canonical polyadic (CP) decomposition) over an aggregation of the outputs of all layers in Equation~\eqref{GNN_MLP_eq} to provide structure-aware representations of the input graph. Then, we obtain the final embedding $\boldsymbol{\mathcal{Z}}$ by combining embeddings from the above modules, i.e., $\boldsymbol{\mathcal{Z}} = [\boldsymbol{\Tilde{\mathcal{Z}}}_{\text{PIT}}, \boldsymbol{\mathcal{Z}}_{\mathcal{G}}]$, where $[\cdot,\cdot]$ denotes the concatenation operation and $\boldsymbol{\mathcal{Z}}_{\mathcal{G}}$ represents the final output of the graph convolutional layer. Finally, we feed the final embedding $Z$ into an MLP layer and use a differentiable classifier (here we use a softmax layer) to make graph classification.

\paragraph{\bf Conditional Conformal Prediction.}
Our proposed {\em conditional uncertainty quantification} algorithm aims at constructing the {\em conditional prediction set} $C^{cond}(\mathcal{G}) $ for any graph $\mathcal{G}\in\mathcal{G}_{test}$ with label $y$, such that for any user-chosen error rate $\alpha\in[0,1]$,
\begin{equation}\label{eq: cov cond}
    \mathbb{P}\big(y_{\text{test}} \in C^{cond}(\mathcal{G}_{\text{test}})\mid \mathcal{G}\big) \geq 1 - \alpha,
\end{equation}
where the probability measure is defined on the augmented data set $\mathcal{G}_{train},\mathcal{G}_{calib}$ and only $y$ with $\mathcal{G}$ fixed (or from some local distribution).

Our idea is to only use the local (or similar) samples to obtain a prediction set. In particular, we define a local calibration set $\mathcal{G}_{\text{calib}}^{\text{local}}$ that contains data with similar graph structure, and assume $\overline{\mathcal{G}} \sim \overline{\mathcal{G}}_{\text{calib}}^{\text{local}} \sim \overline{\mathcal{F}}_{\mathcal{G}}$. Then we can replace $\mathcal{G}_{\text{calib}}$ with $\mathcal{G}_{\text{calib}}^{\text{local}}$ to obtain a conditional prediction set. The detailed algorithm is described in the following.

After obtaining the trained model $\hat f(\cdot)=\hat f(\cdot\mid\mathcal{G}_{train})$, we modify the original calibration steps to construct the conditional prediction set. 
\begin{enumerate}[label=(\arabic*)]
\item Calculate the non-conformity score: 
$$s_i = \mathcal{A}\big((y_i,\mathcal{G}_i),\mathcal{G}_{train}\big)= 1-\hat f(\mathcal{G}_i)_{y_i}, \quad\text{for}\quad\mathcal{G}_i\in \mathcal{G}_\text{calib}^{local},$$
where $\hat f(\mathcal{G}_i)_{y_i}$ is the predicted probability of the $i$-th subject having label $y_i$. Similarly we define $s_{\mathcal{G}}(y) = 1-\hat f(\mathcal{G})_{y}$. 
\item Define a conformal p-value 
$p^{y}(\mathcal{G}) = \frac{\sum_{j:\mathcal{G}_j\in\mathcal{G}_{calib}^{local}} {\bf 1}(s_j\le s_{\mathcal{G}}(y))+1}{|\mathcal{G}_{calib}^{local}|+1}$ that directly indicate a $(1-\alpha)$-level prediction set. 
\item The prediction set can be defined as: 
\begin{equation} \label{eq:CP set uncond}
   C^{\text{cond}}(\mathcal{G}) = \{y : p^{y}(\mathcal{G}) \geq \alpha\}
\end{equation}
\end{enumerate}
We will discuss the similarity measure between tensorized graphs in the sequel. The theoretical coverage is established in Proposition \ref{prop: cond cov}.
\begin{Proposition} \label{prop: cond cov}
    Assume $\mathcal{G}_{calib}^{local}$ and $\mathcal{G}$ are exchangeable and independent with $\mathcal{G}_{train}$, the conformal prediction set $C^{cond}(\mathcal{G})$ obtained in this section satisfies (\ref{eq: cov cond}) with conditional probability measure defined on the augmented data set $\mathcal{G}_{train},\mathcal{G}_{calib}$ and only $y$ with $\mathcal{G}$ from the local distribution as $\mathcal{G}_{calib}^{local}$.
\end{Proposition}

{\it
\noindent
\underline{Proof of Proposition~\ref{prop: cond cov}}:
For notation simplicity, we index $\mathcal{G}_{calib}^{local}$ by $1,\cdots,m$ and the target $(\mathcal{G},y)$ by $m+1$. Let $s_{m+1} = s_{\mathcal{G}}(y)$, $s_{m+1}$ and $s_i$ are i.i.d. given $\mathcal{G}_{train}$.
{
\begin{align*}
    Pr\big(y \in C^{\text{cond}}(\mathcal{G})\mid \mathcal{G}\sim \mathcal{F}_{\mathcal{G}}\big)&= Pr\left(\frac{1}{m+1}\left[1+\sum_{i=1}^{m+1}{\bf 1}(s_{m+1}>s_i)\right]\leq \alpha\right)\\
    &=\mathbb{E} \left[{\bf 1}\left(\frac{1}{m+1}\left[1+\sum_{i=1}^{m+1}{\bf 1}(s_{m+1}\geq s_i)\right]\leq \alpha \right)\right]\\
    &= \frac{1}{m+1} \mathbb{E}\left[\sum_{j=1}^{m+1} {\bf 1}\left(\frac{1}{m+1}\left[1+\sum_{i=1}^{m+1}{\bf 1}(s_j\geq s_i)\right]\leq \alpha\right)\right]
\end{align*}}
Order $\{s_i\}_{i=1}^{m+1}$ by $s_{(1)}<s_{(2)}<\ldots<s_{(m+1)}$, and as $s_i$ are continuous, we ignore the tie here. For $s_j = s_{(i)}$, we have that $\sum_{t=1}^{m+1}{\bf 1}(s_j\ge s_t) = i $.
Given $\alpha\in (0,1)$, there exists $k\in\{0,1,\ldots,m\}$ such that $k<  (m+1)\alpha \le k+1$. Thus
{
$$
\frac{1}{m+1}\sum_{j=0}^{m+1} {\bf 1}\big(\frac{1}{m+1}\big[1+\sum_{i=1}^{m+1}{\bf 1}(s_j> s_i)\leq \alpha   \big)\big] = \frac{k+1}{m+1}\ge\alpha.
$$ 
}
\hfill $\diamond$
}

\paragraph{\bf Graph Similarity Measurement.} For a proper definition of a neighbourhood $\mathcal{N}_{\mathcal{G}}$, the ``similarity'' can be measured in different ways. In this work, we consider two approaches: (i) {\it topological similarity}: Let $Dg(\mathcal{G}_i)$ and $Dg(\mathcal{G}_j)$ be persistence diagrams of two graphs $\mathcal{G}_i$ and $\mathcal{G}_j$, respectively. Then we measure topological similarity among $Dg(\mathcal{G}_i)$ and $Dg(\mathcal{G}_j)$ as $d_{W_p}(\mathcal{G}_i, \mathcal{G}_j) =\inf_{\gamma} \biggl(
\sum_{{ x\in Dg(\mathcal{G}_i) \cup \Delta}}
||x-\gamma(x)||^p_{\infty}
\biggr)^{\frac{1}{p}}
$, where $p \geq 1$ and $\gamma$ is taken over all bijective maps between  $Dg(\mathcal{G}_i) \cup \Delta$ and $Dg(\mathcal{G}_j) \cup \Delta$, counting their multiplicities. In our analysis we use $p=1$; and (ii) {\it similarity in the learned
embedding space}: We measure the graph similarity based on the similarity between graph-level embeddings from our CF-T$^2$NN model. That is, the similarity between the two graphs is provided by their Euclidean distance in the CF-T$^2$NN embedding space. We report classification performances of CF-T$^2$NN under the above two graph similarity measures (see Table~\ref{classification_result_0_bio_graphs}). Figure~\ref{fig:flowchart} illustrates our proposed CF-T$^2$NN framework.
\section{Experiments}
The evaluation results on 10 datasets and the ablation study on 6 datasets are summarized in Tables~\ref{classification_result_0_bio_graphs} and~\ref{ablation_study} respectively.
\subsection{Experiment Settings}
\paragraph{\bf Datasets.}
We validate the Uncertainty Quantification (UQ) performance of the CF-T$^2$NN model on graph classification tasks across two main types of datasets, chemical compounds and protein molecules. For the chemical compound dataset, we utilize four datasets, namely MUTAG, DHFR, BZR, and COX2~\cite{sutherland2003spline,kriege2012subgraph}, which comprise graphs representing chemical compounds, where nodes denote different atoms, and edges represent chemical bonds. For the molecular compound datasets, six datasets are used, including ENZYMES, PROTEINS, PTC\_MR, PTC\_MM, PTC\_FM, and PTC\_FR~\cite{helma2001predictive,schomburg2004brenda,dobson2003distinguishing,borgwardt2005protein,kriege2012subgraph}. In molecular compound datasets, nodes represent amino acids, while edges denote the relationships or interactions between amino acids, such as physical bonds, spatial proximity, or functional interactions. For all graphs datasets, we adopt a split ratio of 0.5/0.2/0.21/0.09 for training, testing, calibration, and validation subsets, respectively. Table~\ref{tab:datasets} summarizes the characteristics of all ten datasets used in our experiments.
\begin{table}[h!]
\centering
\caption{Summary statistics of the benchmark datasets.\label{tab:datasets}}
\resizebox{0.55\columnwidth}{!}{
\begin{tabular}{lcccc}
\toprule
\textbf{{Dataset}} & \textbf{{\# Graphs}} &\textbf{{Avg.} $|\mathcal{V}|$} & \textbf{{Avg.} $|\mathcal{E}|$} & \textbf{{\# Class}} \\
\midrule
ENZYMES & 600 & 32.63 & 62.14 & 6\\
BZR &405 &35.75 &38.35 &2 \\
COX2 &467 &41.22 &43.45 &2 \\
DHFR & 756 & 42.43& 44.54 & 2\\
MUTAG &188 &17.93 &19.79 &2 \\
PROTEINS &1113 &39.06 &72.82 &2 \\
PTC\_MR &  344 & 14.29 & 14.69 & 2\\
PTC\_MM &  336 & 13.97 & 14.32 & 2\\
PTC\_FM &  349 & 14.11 & 14.48 & 2\\
PTC\_FR & 351 & 14.56 & 15.00 & 2 \\
\bottomrule
\end{tabular}}
\end{table}

\noindent{\bf Baselines.}
We compare our CF-T$^2$NN with 7 state-of-the-art (SOTA) baselines including (i) Graph Convolutional Network (GCN)~\cite{kipf2016semi}, (ii) Chebyshev GCN (ChebNet)~\cite{defferrard2016convolutional}, (iii) Graph Isomorphism Network (GIN)~\cite{xu2018powerful}, (iv) GNNs with Differentiable Pooling (DiffPool)~\cite{ying2018hierarchical}, (v) Self-attention Graph Pooling (SAGPool)~\cite{lee2019self}, (vi) Topological Graph Neural Networks (GNN-TOGL)~\cite{horntopological}, and (vii) Simplicial Isomorphism Networ (SIN)~\cite{bodnar2021weisfeiler}.

\paragraph{\bf Experimental Setup.}
We run our experiments on a single NVIDIA Quadro RTX 8000 GPU card, which has up to 48GB of memory. To train the end-to-end CF-T$^2$NN model, we use the Adam optimizer with a learning rate of 0.001.  We use ReLU as the activation function $\sigma(\cdot)$ across our CF-T$^2$NN model, except Softmax for the MLP classifier output. For the resolution of the $\text{PI}_{\text{Dg}}$, we set the size to be $P = 50$. In our experiments, we consider $\mathcal{K} = 4$ different filtrations, i.e., degree-based, betweenness-based, closeness-based, and eigenvector-based filtrations. Depending on the dataset, we set batch sizes of either 16 or 32. The optimal number of hidden units for each layer in the graph convolution and MLPs are explored from the search space $\{16, 32, 64, 128, 256\}$. The number of hidden units of TTL is 32. In addition, our CF-T$^2$NN model has 3 layers in the graph convolution blocks and 2 layers in the MLPs, with a dropout rate of 0.5 for all datasets. We train our model for up to 100 epochs to make sure it is fully learned. To conduct the conditional conformal prediction, the number of neighbors of the $k$-nearest neighbor method is set to either 80 or 100 (depending on the dataset), with a user-chosen error rate $\alpha$ of 0.1. Our datasets and codes are available on~\url{https://www.dropbox.com/scl/fo/bi6x0hm83goxe7o0he5yr/h?rlkey=11sbequeppet9zp39blohjepa&dl=0}. 

\subsection{Results}
Table~\ref{classification_result_0_bio_graphs} shows the comparison of our proposed CF-T$^2$NN with state-of-the-art baselines for graph classification on 10 datasets. We consider two variants of CF-T$^2$NN, i.e., CF-T$^2$NN$_T$ (i.e., based on topological similarity) and CF-T$^2$NN$_E$ (i.e., based on similarity in the learned embedding space). The results indicate that our CF-T$^2$NN always achieves the best performance, i.e., the average size of the prediction sets. Particularly, from Table~\ref{classification_result_0_bio_graphs}, we observe that: (i) Compared to the GCN-based models (i.e., GCN and ChebNet), CF-T$^2$NN improves upon the runner-up (i.e., ChebNet) by a margin of 19.12\%, 26.02\%, 36.72\%, and 7.69\% on COX2, DHFR, BZR, and MUTAG datasets respectively; (ii) Compared to the graph pooling methods (i.e., DiffPool and SAGPool), CF-T$^2$NN yields an average relative gain of 13.36\% on all datasets; and (iii) Compared to topology-based deep learning models (i.e., SIN and TOGL), CF-T$^2$NN outperforms both models with a significant margin, especially on chemical graphs.

\begin{table}[h!]
\centering
\caption{Performance on molecular and chemical graphs. The best results are given in {\bf bold}. 
\label{classification_result_0_bio_graphs}}
\resizebox{1.\columnwidth}{!}{
\begin{tabular}{lcccccccccc}
\toprule
\textbf{{Model}} &\textbf{{ENZYMES}} & \textbf{{COX2}} & \textbf{{DHFR}}  & \textbf{BZR} & \textbf{{MUTAG}} & \textbf{{PROTEINS}} &\textbf{{PTC\_MR}} & \textbf{PTC\_MM} & \textbf{PTC\_FM} & \textbf{PTC\_FR}\\
\midrule
GCN~\cite{kipf2016semi} & 5.57$\pm$0.36 & 1.64$\pm$0.58 & 1.55$\pm$0.62 & 1.77$\pm$0.59 & 1.56$\pm$0.42 & 1.99$\pm$0.58 & 1.85$\pm$0.35 & 1.56$\pm$0.47 & 1.87$\pm$0.38 & 1.78$\pm$0.30\\
ChebNet~\cite{defferrard2016convolutional} & 5.64$\pm$0.29 &  1.62$\pm$0.48 & 1.55$\pm$0.64 & 1.75$\pm$0.55 & 1.54$\pm$0.49 & 1.96$\pm$0.57 & 1.87$\pm$0.36 & 1.59$\pm$0.50 & 1.85$\pm$0.32 & 1.71$\pm$0.35\\
GIN~\cite{xu2018powerful} & 5.38$\pm$0.19& 1.53$\pm$0.49&1.36$\pm$0.71 & 1.70$\pm$0.50&1.56$\pm$0.42 &1.93$\pm$0.59 &1.78$\pm$0.31&1.48$\pm$0.44 & 1.81$\pm$0.29 &1.70$\pm$0.36\\
DiffPool~\cite{ying2018hierarchical} & 5.21$\pm$0.24 & 1.65$\pm$0.47 & 1.47$\pm$0.57 & 1.70$\pm$0.40 & 1.58$\pm$0.52 & 1.91$\pm$0.48 & 1.83$\pm$0.35 & 1.48$\pm$0.49 & 1.86$\pm$0.31 & 1.72$\pm$0.44\\
SAGPool~\cite{lee2019self} & 5.27$\pm$0.33 & 1.57$\pm$0.40 & 1.39$\pm$0.66 & 1.72$\pm$0.59 & 1.55$\pm$0.41 & 1.85$\pm$0.41 & 1.79$\pm$0.32 & 1.49$\pm$0.52 & 1.81$\pm$0.46 & 1.76$\pm$0.36\\
SIN~\cite{bodnar2021weisfeiler} & 5.73$\pm$0.32 & 1.50$\pm$0.43 & 1.44$\pm$0.77 & 1.64$\pm$0.56 & 1.53$\pm$0.57 & 1.69$\pm$0.43 & 1.80$\pm$0.35 & 1.49$\pm$0.45 & 1.85$\pm$0.32 & 1.72$\pm$0.31\\
TOGL~\cite{horntopological} & 5.06$\pm$0.25 & 1.51$\pm$0.58 & 1.36$\pm$0.74 & 1.65$\pm$0.50 & 1.49$\pm$0.45 & 1.58$\pm$0.47 & 1.86$\pm$0.37 & 1.52$\pm$0.45 & 1.82$\pm$0.40 & 1.74$\pm$0.30\\
\midrule
\textbf{CF-T$^2$NN$_T$ (ours)} & 4.48$\pm$0.38&1.48$\pm$0.45 &{\bf 1.23$\pm$0.49} & 1.54$\pm$0.43&{\bf 1.43$\pm$0.50} &{\bf 1.55$\pm$0.42}&1.75$\pm$0.33 &{\bf 1.45$\pm$0.44} &1.84$\pm$0.27 &1.69$\pm$0.36\\
\textbf{CF-T$^2$NN$_E$ (ours)} & {\bf 4.61$\pm$0.82}&{\bf 1.36$\pm$0.47} &{1.28$\pm$0.48} &{\bf 1.28$\pm$0.48} &1.44$\pm$0.50 &1.56$\pm$0.41& {\bf 1.69$\pm$0.36}&1.45$\pm$0.45 &{\bf 1.72$\pm$0.35} & {\bf 1.66$\pm$0.38}\\
\bottomrule
\end{tabular}}
\end{table}
\begin{table}[h!]
\centering
\caption{Ablation study of the CF-T$^2$NN.\label{ablation_study}}
\resizebox{1.\columnwidth}{!}{
\begin{tabular}{lcccccccccc}
\toprule
\textbf{{Architecture}} & \textbf{{ENZYMES}} & \textbf{{COX2}} & \textbf{{MUTAG}} &\textbf{{PTC\_MR}} & \textbf{PTC\_MM} &  \textbf{PTC\_FR}\\
\midrule
CF-T$^2$NN w/o TTL &4.53$\pm$0.43 & 1.37$\pm$0.48&  1.48$\pm$0.48 & 1.78$\pm$0.32 & 1.48$\pm$0.48 & 1.81$\pm$0.29\\
CF-T$^2$NN & {\bf 4.48$\pm$0.38}& {\bf 1.36$\pm$0.47}& {\bf 1.43$\pm$0.50}&{\bf 1.69$\pm$0.36} & {\bf 1.45$\pm$0.44} &{\bf 1.66$\pm$0.38} \\
\bottomrule
\end{tabular}}
\end{table}

\subsection{Ablation Study}
To better understand the importance of the tensor transformation layer (TTL) and the impact of uncertainty quantification (UQ) on performance across different architectures, we design ablation study experiments on the ENZYMES, COX2, MUTAG, PTC\_MR, PTC\_MM, and PTC\_FR datasets. As shown in Table~\ref{ablation_study}, CF-T$^2$NN without the TTL module results in an average performance decline of over 5.04\%, which highlights the critical role of learning the tensor structure of features. This ablation of TTL components not only leads to a larger prediction set size across all datasets but also underscores the efficacy of our uncertainty quantification method. Despite the performance drop, our UQ approach, particularly through conformal prediction, maintains strong statistical coverage of the prediction set results, ensuring reliability even in the absence of TTL. In summary, the ablation study not only confirms the integral role of TTL in the CF-T$^2$NN model but also showcases the resilience of our UQ measures in providing dependable predictions.


\section{Conclusion}

In conclusion, our research has successfully addressed a significant gap in the field of GNNs by developing a novel model that robustly quantifies uncertainty, even in the presence of covariate shifts within graph-structured data. By integrating tensorized topological neural networks with an extended conformal prediction framework, we have introduced a method that not only adheres to statistical guarantees of coverage for conditional prediction sets but also respects the complex, multi-dimensional nature of graph data. This approach has proven effective across a diverse set of real-world datasets, showcasing its superiority over existing state-of-the-art methods in terms of reliability and precision. Our findings underscore the importance of considering domain shifts and the high-dimensional challenges inherent in graph data when quantifying uncertainty. This work sets a new standard for the application of GNNs in critical decision-making scenarios, offering a significant contribution to the fields of machine learning and graph analysis. We believe that our methodology will inspire further research into robust uncertainty quantification for GNNs and encourage the development of more reliable and interpretable models for complex systems analysis.

\clearpage

\bibliography{bib/conformal,bib/tensor_network,bib/neurips_2024,bib/uqgnn,bib/ec}
\end{document}